\documentclass{article}
\usepackage[british]{babel}
\usepackage{jfrExamplee}
\usepackage{graphicx}
\usepackage{setspace}

\usepackage[style=apa,backend=biber,bibencoding=utf8]{biblatex}
\DeclareLanguageMapping{british}{british-apa}
\AtEveryBibitem{\clearfield{month}}
\AtEveryBibitem{\clearfield{day}}

\usepackage{gensymb}
\usepackage{amsmath}
\usepackage{xcolor}
\usepackage{multicol}

\usepackage{endnotes}
\let\footnote=\endnote

\title{Design and Evaluation of a Modular Robotic Plum Harvesting System Utilising Soft Components}

\author{
	Jasper Brown \\
	The Australian Center for Field Robotics \\
	The University of Sydney \\
	Sydney, Australia \\
	\texttt{j.brown@acfr.usyd.edu.au} \\
	\And
	Salah Sukkarieh \\
	The Australian Center for Field Robotics \\
	The University of Sydney \\
	Sydney, Australia \\
	\texttt{salah@acfr.usyd.edu.au}
}

\begin{document}
	\maketitle
	
	\begin{abstract}
		The human labour required for tree crop harvesting is a major cost component in fruit production and is increasing. To address this, many existing research works have sought to demonstrate commercially viable robotic harvesting for tree crops, though successful commercial products resulting from these have been few and far between. Systems developed for specific crops such as sweet peppers or apples have shown promise, but the vast majority of cultivar types remain unaddressed and developing a specific system for each one is inefficient. In this work a flexible and modular development platform is presented, this can be used to test specific design choices on different fruit and growing conditions. The system is evaluated in a commercial plum orchard, with no crop modifications. Some existing techniques are found to be counterproductive for plums, while soft robotics and persistent target tracking significantly improve performance. A harvest success rate of 42\% was observed, with lower than expected effectiveness based on prior testing with apples. The plum type and growing style pose unique challenges which are examined in the context of system module design choices. 
		
		
	\end{abstract}
	
	\section{Introduction}
Robotic fruit harvesting is a problem that has captured the attention of researchers for over 40 years, who have generated a wealth of publications on the topic. The economic benefits to growers, of reduced reliance on a largely seasonal and often untrained labour force are clear. Despite an obvious motivation and extensive research, translating research outcomes into commercial solutions has proven difficult. Those which have recently emerged are limited to specific crop types and the vast majority of tree crop cultivars, such as plums, remain unaddressed. What we present here is a system design that uses modular software with highly flexible hardware to allow for rapid development of system components that may be specific to different growing conditions or fruit cultivars. By using this flexible platform for development, we are able to identify key modules which remain common between problems, and provide minimum hardware requirements for each use case. This system is validated on a plum crop and studies of the design decisions for various components are carried out. In addition to commonly used techniques, a persistent target detection and filtering module is introduced to overcome the limitations of object detection in highly obscured crops. Also, the use of soft robotics components in the gripper hardware module was shown to have a very large impact on harvest success while eliminating issues associated with collisions and tree damage.

The primary limitation of this work is the small amount of testing data, which is restricted to a single growing season. We plan to address this in the coming year. Gripper force and longevity were also an issue, with further design iterations of the soft gripper geometry and materials required. Assessment of non-pickable fruit to avoid, and autonomous picking motion selection, are identified as areas for possible system performance improvement.

\section{Related Work}
There is an extensive body of literature on both systems and techniques for robotic fruit picking. Over 50 papers have been published on the topic, many of these in the past several years~\parencite{bac2014, comba2010, de_an2011, }. Despite significant research effort, commercial progress has been slow and the field has been hampered by the difficulty of implementing standard testing protocols to allow direct comparison. Many works are specific to one crop type and growing setup, while the difficulty of creating common testing setups makes direct comparisons largely impossible even within a fruit type.  

System implementations of robotic harvesters using modern techniques are presented by~\textcite{arad2020} for sweet peppers and by~\textcite{xiong2019} for strawberries. Many additional research efforts have been directed towards autonomous harvesting in protected cultivation systems~\parencite{van_henten2002,van_henten2013, tanigaki2008,}. The growing pattern and lighting regularity of these environments, along with typically higher value crops, makes them ideal candidates for automated harvesting. Crops such as sweet peppers possess soft and flexible stems, in contrast to the hard lignified wood of tree crops, meaning collisions will easily damage the plant but will not often lead to stoppage of the harvesting process. 

Operating on outdoor tree crops brings the opposite challenge, with system component damage more likely than tree damage. It also introduces additional complexities in perception, gripping and crop variation. Apple harvesting with a low cost 3D printed gripper was demonstrated by~\textcite{silwal2017} but occurred on a modified crop. A custom designed manipulator arm with a prismatic base is used to reduce cost and simplify planning. Rotational motion is applied during picking. Common failure causes were identified as position error, followed by apples on long shoots which behave as pendulums when picked. 

The~\textcite{silwal2017} prototype was iterated upon in~\textcite{hohimer2019} which performed apple harvesting on an unmodified crop using a soft pneumatic gripper. They identify fruit clustering as the most common failure case, followed by positioning error. Multiple exposures are used to increase image dynamic range and an ensemble of two learned feature detectors are applied to the resulting image. Picking motion is a straight pull and operating with a non zero pitch angle was not found to improve performance, but did increase collisions. No research activity on robotic plum harvesting was found during a review of the literature, however mechanised plum harvesting is assessed in~\textcite{mika2015}.

Commercial systems are beginning to become available for crops such as tomatoes (Panasonic\footnote{news.panasonic.com/global/stories/2018/57801.html. Accessed on 28/4/2020}), strawberries (Agrobot\footnote{agrobot.com. Accessed on 27/4/2020}, Octinion\footnote{octinion.com/products/agricultural-robotics/rubion. Accessed on 28/4/2020}, CROO Robotics\footnote{harvestcroo.com. Accessed on 28/4/2020}), raspberries (fieldwork robotics\footnote{phys.org/news/2019-05-fieldwork-robotics-field-trials-raspberry.html. Accessed on 28/4/2020}) and apples (Abundant Robotics\footnote{abundantrobotics.com. Accessed on 28/4/2020}, FFRobotics\footnote{ffrobotics.com. Accessed on 28/4/2020}) among others. As of writing, all systems targeted at tree crops remain in the development phase. These solutions target a single type of fruit and require specific growing styles to be effective. Developing harvesting robots for less common fruit and growing systems will likely remain an open problem for many years and creates the need for flexible development platforms such that proposed in this work. 

Fruit detection is the first step in picking and has been done using both hand engineered and learned approaches~\parencite{kapach2012, vitzrabin2016b}. Dynamic thresholding in a range of colour spaces is applied to detect common crops in \textcite{zemmour2019}. They report comparable results to deep learning approaches, though with minimal training data. Multi-spectral sensing has also been explored as an alternative to RGB colour channels for fruit detection~\parencite{hung2013}. Synthetic image generation is explored by~\textcite{barth2018} as a means of overcoming the need to gather large training datasets. A deep network is combined with morphological \textcite and colour thresholding to yield high frame rates while detecting sweet peppers and fruit in~\textcite{arad2020}, though this operates on yellow fruit. Lighting control is effective for greenhouse environments through filtering and flash-no-flash image sequences as demonstrated in~\textcite{arad2019}. State of the art fruit counting results are generated in~\textcite{chen2017} using a multi-stage deep learning approach optimised for overall fruit quantity estimation, rather than individual instance localisation. Multi-view detection has also been considered~\parencite{hemming2014}. 

Target tracking and information fusion over multiple frames is rarely implemented in harvesting system designs. In work by~\textcite{mehta2017} multiple monocular cameras are used with a particle filtering framework to localise fruit in 2D. Spring-mass motion models are developed, though only validated in simulation. Disparity maps are used to detect and measure broccoli seedlings in~\textcite{ge2019}. Outliers are removed by applying K-nearest neighbours to the point clouds but no filtering occurs on the final detections.

Gripper design is a critical component of autonomous harvesting~\parencite{rodriguez2013}. In addition to the pneumatic gripper of~\textcite{hohimer2019}, other experimental techniques include under-actuated cable grippers~\parencite{xiong2019} and integrating tactile feedback~\parencite{dimeas2015}. The Abundant Robotics commercial platform uses a vacuum type gripper that swallows the fruit. A commercial deformable-finger end effector design is used by~\textcite{eizicovits2016}, where simulation tools are used to map grasp position tolerances to sensing requirements. Crops such as sweet peppers require stem cutting which has been addressed in combination with suction~\parencite{bac2017, lehnert2017} and catching~\parencite{arad2020} type grippers. 

Final target approach is commonly done using Image Based Visual Servoing (IBVS) with eye-in-hand cameras \parencite{arad2020, mehta2016, barth2016}. A review of vision based control is presented in~\textcite{zhao2016}. End effector integrated IR distance sensors are used by~\textcite{xiong2019} for final positioning. Multiple sweet pepper approach strategies are assessed under laboratory and greenhouse conditions by~\textcite{ringdahl2019a}. Using multiple approach attempts was found to slightly increase the probability of reaching a pepper, at the cost of higher cycle times. Radicchio harvesting is targeted in~\textcite{foglia2006} which requires cutting the plant stem approximately 10mm below the ground plane using a claw type mechanism. High density 2D growing systems are well suited to cartesian picking motions with up to 91\% of fruit reachable in this manner~\parencite{vougioukas2016}.

In this work two new gripper designs featuring both hard and soft components are compared. Additionally, a persistent target tracking filter is developed to go beyond single-frame perception techniques. Unlike existing works, significant emphasis is placed on the modularity of hardware and software. This modular approach is tested by harvesting a novel crop type, with multiple module design choices being evaluated.

\section{Experimental Design}
Full system evaluation was carried out as part of a week long plum harvesting trial. The aim of this is was to discover systemic strengths and weaknesses of the modular framework, and to understand the requirements specific to plum crop harvesting. Within this trial, three experiments were done to test specific module design choices. In the first, multiple object detector algorithms are tested in both day and night scenes. Next, both simple and complex harvesting motions are assessed for a soft gripper. Finally, hard and soft grippers are compared on criteria of their effectiveness and robustness to collisions.

\begin{figure}[h!]
	\centering
	\includegraphics[width=0.6\columnwidth]{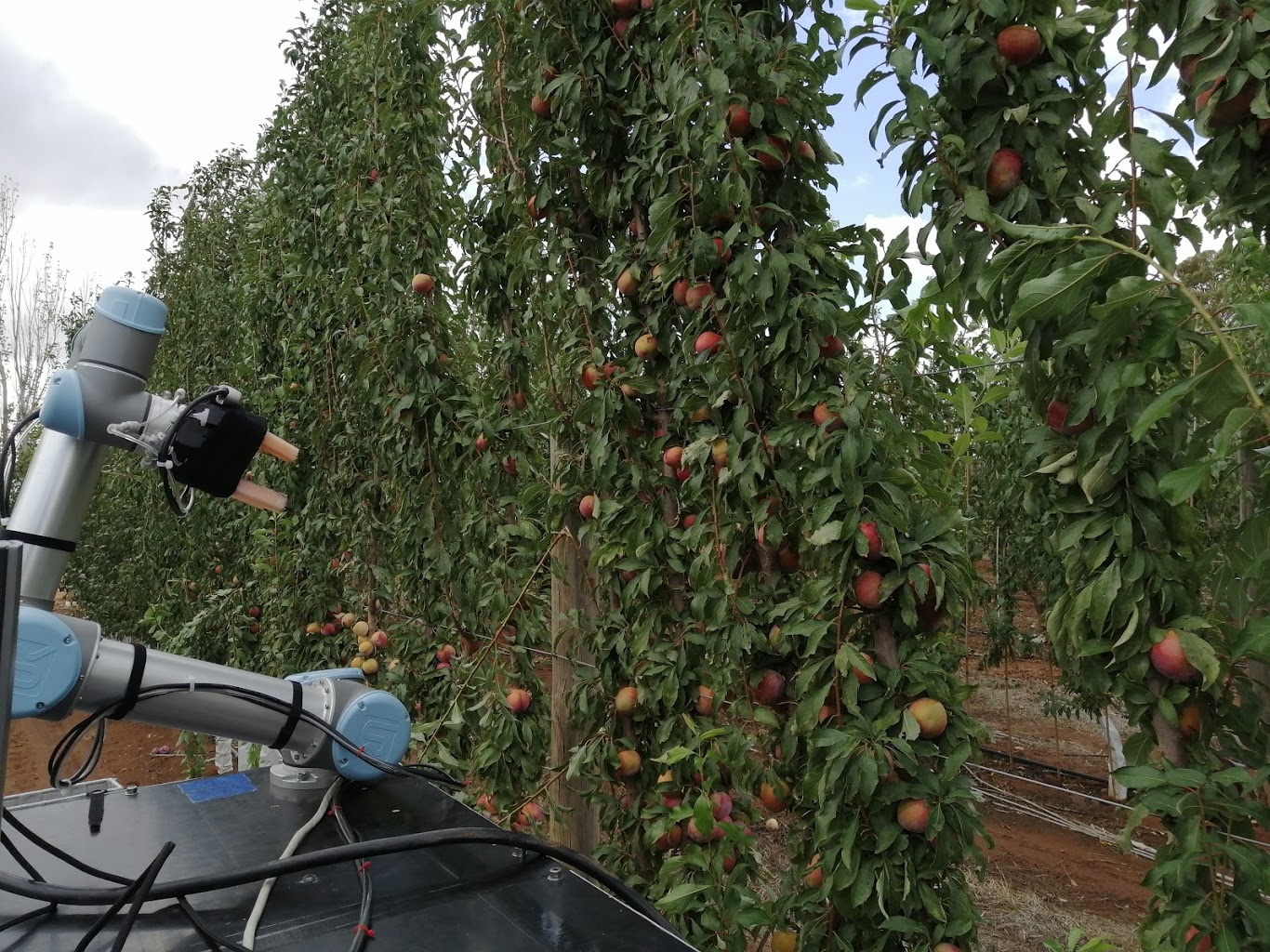}
	\caption{The system set up in front of a representative 2D trellis panel. The plum trees are trained by splitting the base into 4 fruit bearing vertical trunks which are fixed to the trellis system and pruned yearly into to a flat plane.}
	\label{fig:trellis}
\end{figure}

Target crop and trellis style are essential considerations for autonomous harvesting. Plums were chosen for this work due to their availability in a modern fruiting wall style 2D trellis configuration. This 2D style, shown in~Figure \ref{fig:trellis}, is widely employed and well suited to mechanisation. It provides for very uniform growing conditions, leading to more predictable and profitable crops. The fruiting wall trellis also provides clear access to the target fruit and is ideal for testing robotic harvesters.

Flower thinning directly determines where fruit will grow in the coming months, so is essential to harvest success. Current best practice thinning, which discourages bunching and matches fruit density to branch carrying capacity, was applied to this crop. A mechanised brush first reduces overall flower load, followed by a manual thinning step.

	\section{Hardware System Design}
\begin{figure}[h!]
	\centering
	\includegraphics[width=0.6\columnwidth]{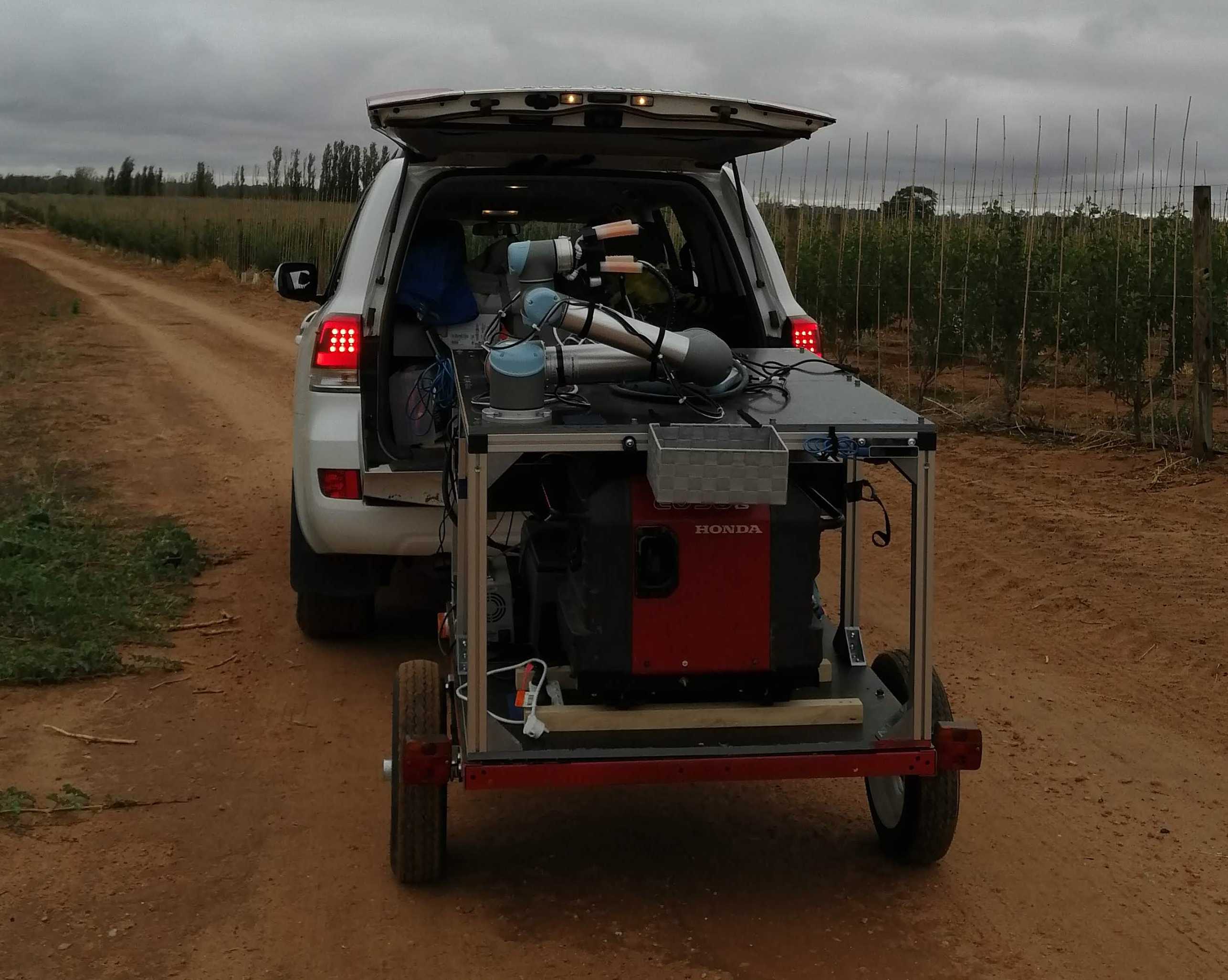}
	\caption{The platform being moved into the field. The arm is stowed and unpowered during transport, display screens are also removed.}
	\label{fig:trailer}
\end{figure}

The hardware system design goals are to realise a fully self contained development platform that has long endurance, is easily modified and can support a variety of payloads. To maximise the deployment options it is constructed on a trailer base, shown in Figure~\ref{fig:trailer}. This can be towed by a robotic platform, moved by hand or attached to existing farm vehicles. The requirements of each hardware module are described here, along with our implementation of these.

\begin{figure}[h!]
	\centering
	\includegraphics[width=0.75\columnwidth]{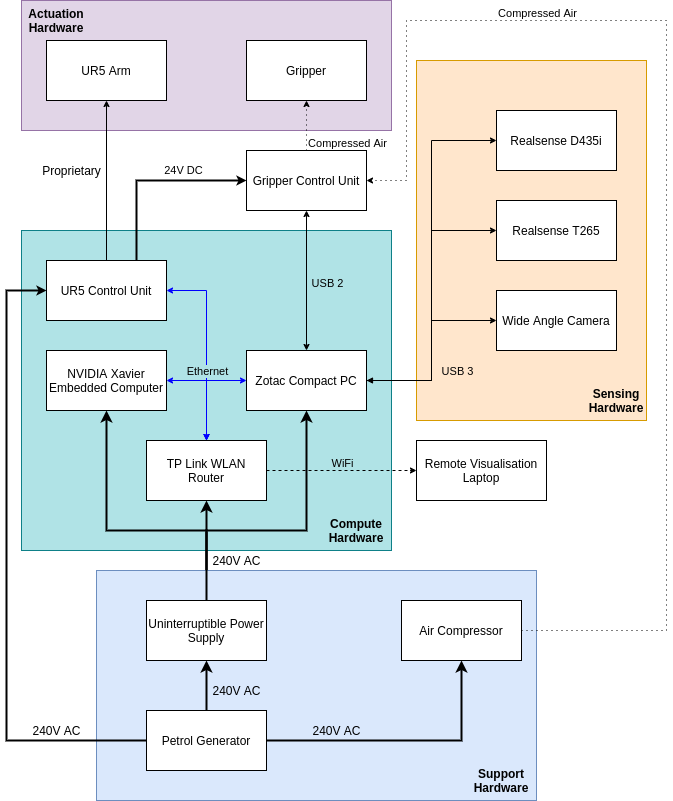}
	\caption{Key hardware modules and connections}
	\label{fig:hw}
\end{figure}

\subsection{Support}
The support hardware module provides 240V power and compressed air at 60psi for the soft gripper. An uninterruptible power supply is also included to provide approximately 30 minutes of battery run time for the compute system while the generator is being refuelled. A system power budget is shown in Table~\ref{table:power} which is met by a 3kW petrol generator. 

\begin{table}[h]
	\centering
		\begin{tabular}{lccc}
			\textbf{Component}                                          & \textbf{\begin{tabular}[c]{@{}c@{}}Nominal \\ Power (W)\end{tabular}} & \textbf{\begin{tabular}[c]{@{}c@{}}Maximum\\ Power (W)\end{tabular}} & \textbf{\begin{tabular}[c]{@{}c@{}}Input \\ Voltage (V)\end{tabular}} \\ \hline
			Compressor                                                  & 50                                                                    & 1000                                                                 & 240                                                                   \\
			Zotac                                                       & 200                                                                   & 330                                                                  & 240                                                                   \\
			Arm                                                         & 150                                                                   & 325                                                                  & 240                                                                   \\
			Xavier                                                      & 40                                                                    & 75                                                                   & 9-20                                                                  \\
			Router                                                      & 10                                                                    & 10                                                                   & 12                                                                    \\
			Arduino                                                     & 1                                                                     & 5                                                                    & 5                                                                     \\
			D435                                                        & 1                                                                     & 2.5                                                                  & 5                                                                     \\
			T265                                                        & 2                                                                     & 2.5                                                                  & 5                                                                     \\
			\begin{tabular}[c]{@{}l@{}}Wide Angle\\ Camera\end{tabular} & 2                                                                     & 2                                                                    & 5                                                                    
		\end{tabular}
	\caption{System power budget. Compressor duty cycle is approximately 5\% resulting in a low average, but high maximum power draw.}
	\label{table:power}
\end{table}

\subsection{Compute}
The compute module provides a high performance PC for generic tasks and an embedded deep learning computer to offload model inference. Low level control of the gripper positioning system, a UR5 manipulator arm, is handled by a dedicated controller, in this case the standard Universal Robots (UR) control unit. An off the shelf router provides local networking as well as a WiFi link for remote control, monitoring and visualisation.

\begin{figure}[h!]
	\centering
	\includegraphics[width=0.7\textwidth]{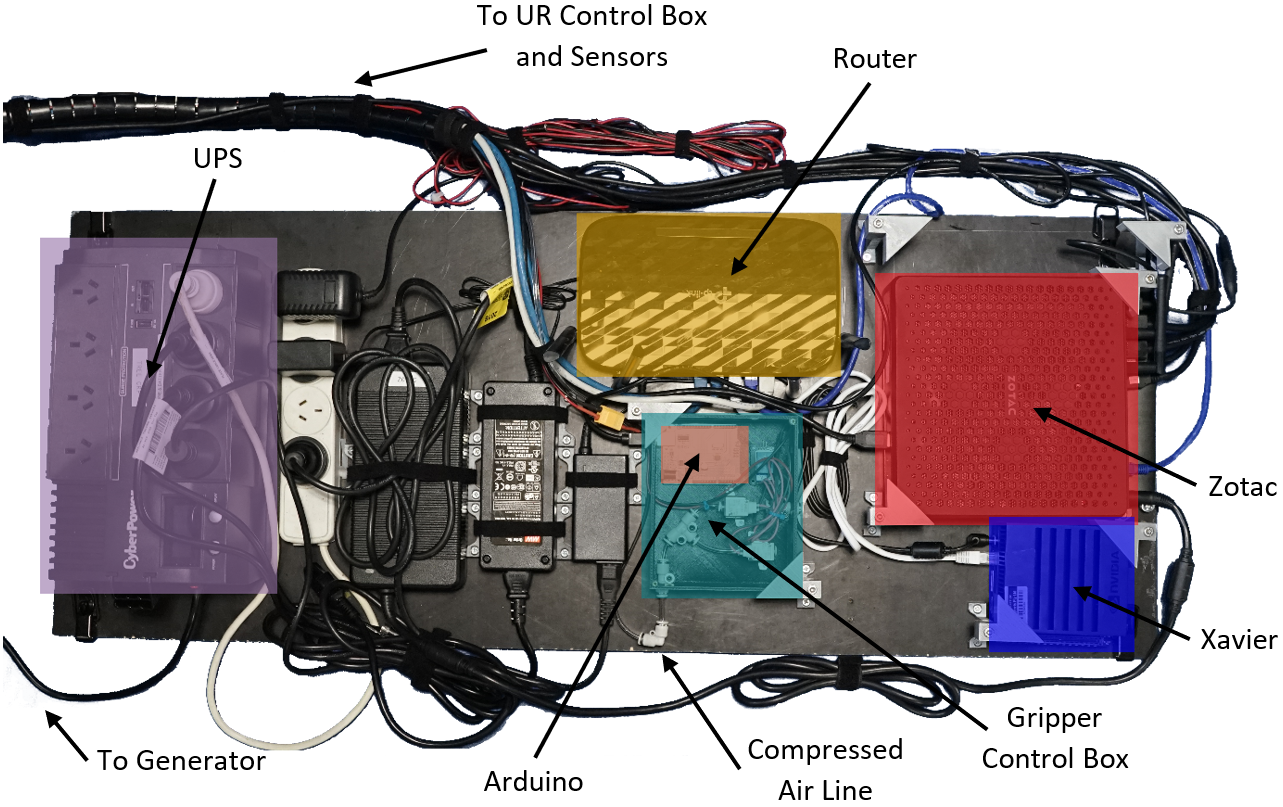}
	\caption{The compute system tray which is self contained and removable from the mobile platform. The wiring harness includes 24V power from the arm, dual HDMI outputs for remote monitoring, 3 camera inputs, air lines from the compressor and to the gripper, also Ethernet and USB connections.}
	\label{fig:compute}
\end{figure}

For the high performance PC, a Zotac EN72080V mini PC is used, this has an i7 processor, 32GB of RAM and an NVIDIA RTX2080 GPU. However, this GPU is only used for development and during deployment the object detection models are run on an NVIDIA Xavier embedded deep learning computer. 

\subsection{Sensing}
Two Intel Realsense cameras are used for 3D sensing, along with a wide angle camera to implement the visual servoing and in-gripper visualisation. A D435i RGBD camera is mounted above the gripper to provide 3D fruit locations and can also be used for obstacle detection. A T265 Simultaneous Localisation and Mapping (SLAM) camera is mounted to the rear corner of the trailer to allow for persistent target tracking as the trailer moves through the orchard. 

The small footprint of the D435i is essential when using the on-gripper mounting configuration and was one of the selection factors when choosing a 3D camera. Many existing approaches use an off-arm camera setup or multiple cameras. The 3D camera is mounted on the end effector to allow for maximum flexibility when positioning it, this can also be used to simulate an off-arm setup by moving the end effector to a fixed position for each new camera frame.

While the D435i provides general fruit localisation, the field of view is insufficient for final approach. Most on-gripper sensing will be obscured at very close range, so a 170\degree diagonal field of view RGB camera is embedded within the soft gripper cup for visual servoing control and grasp visualisation. 

\subsection{Actuation}
Cobot arms have distinct safety advantages, particularly when working in close proximity to untrained farm staff and during platform development. The UR5 CB3 arm was chosen as a well supported and kinematically suitable cobot arm. Workspace and payload limits of the UR5 were sufficient for our development goals. 

Gripper design is an essential aspect of successful fruit harvesting and a huge variety of approaches have been proposed. Additionally, the size, cultivar type and growing conditions of fruit will all alter the optimal gripper design. A specialised gripper control unit is used to provide a standardised software and hardware interface to two grippers tested in this work. Both of these take identical serial commands over USB and are supplied by 24V DC, this commonality allows for rapid integration of existing and future designs.  

\begin{figure}[h]
	\centering
	\includegraphics[width=0.6\columnwidth]{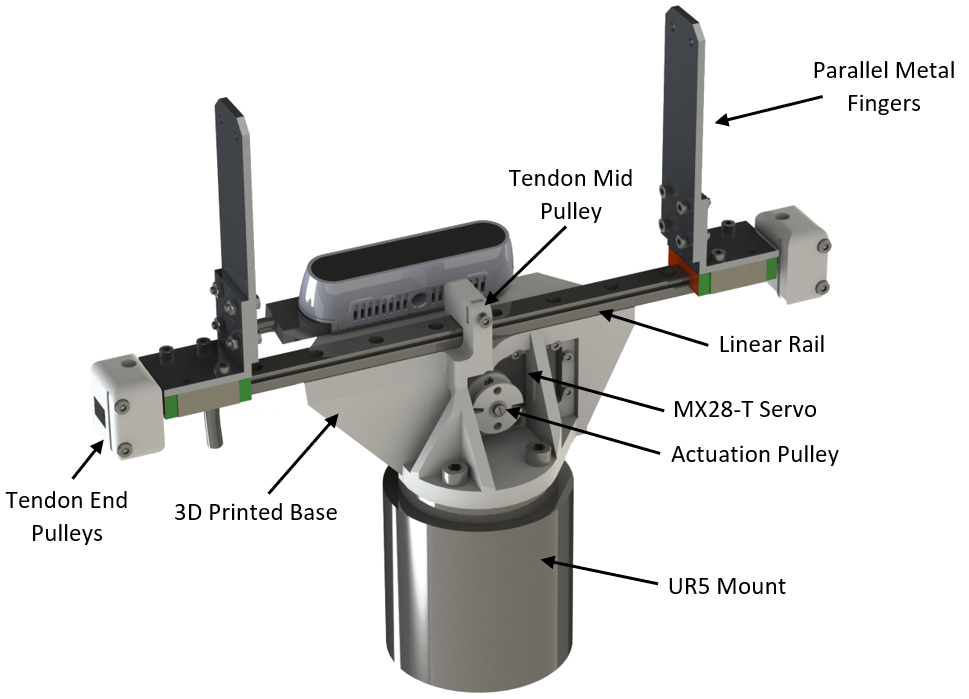}
	\caption{The parallel tendon drive gripper and D435i Realsense. Not rendered are the flexible tendons that run from each sliding car through the mid pulley and onto the actuation pulley. These form a closed loop with the tendon end pulleys to keep constant tension in the drive system as the cars slide along the linear rail.}
	\label{fig:hardGripper}
\end{figure}

A tendon driven parallel gripper, shown in Figure~\ref{fig:hardGripper}, was first developed and is actuated by a Dynamixel Mx28T servomotor, controlled using the OpenCM9.04 board. Two parallel plates are mounted on a linear rail with a common tendon-cam design actuating them simultaneously. By keeping the total tendon loop length constant at all gripper positions, the need for a tensioning mechanism is eliminated allowing for a very simple and low cost gripper that can be made arbitrarily wide. The tested configuration has a minimum closure distance of 15mm and maximum of 200mm. All the components for this are commercially available or 3D printed, and the designs are made available online\footnote{github.com/jaspereb/SimpleSliderHand}.


Following preliminary testing on apple crops, the most common failure mode of the system was identified as fruit being grasped with one side of the parallel gripper fingers on the outside of a branch, leading to the fruit not being held on that side. The most obvious post-grasp mechanism to deal with this was identified as having that finger slide off the branch and snap back into contact with the fruit as the arm retracts. Initial design concepts looked at mechanical linkages, however soft robotics was chosen as a approachable and robust solution. 

\begin{figure}[h]
	\centering
	\includegraphics[width=0.98\columnwidth]{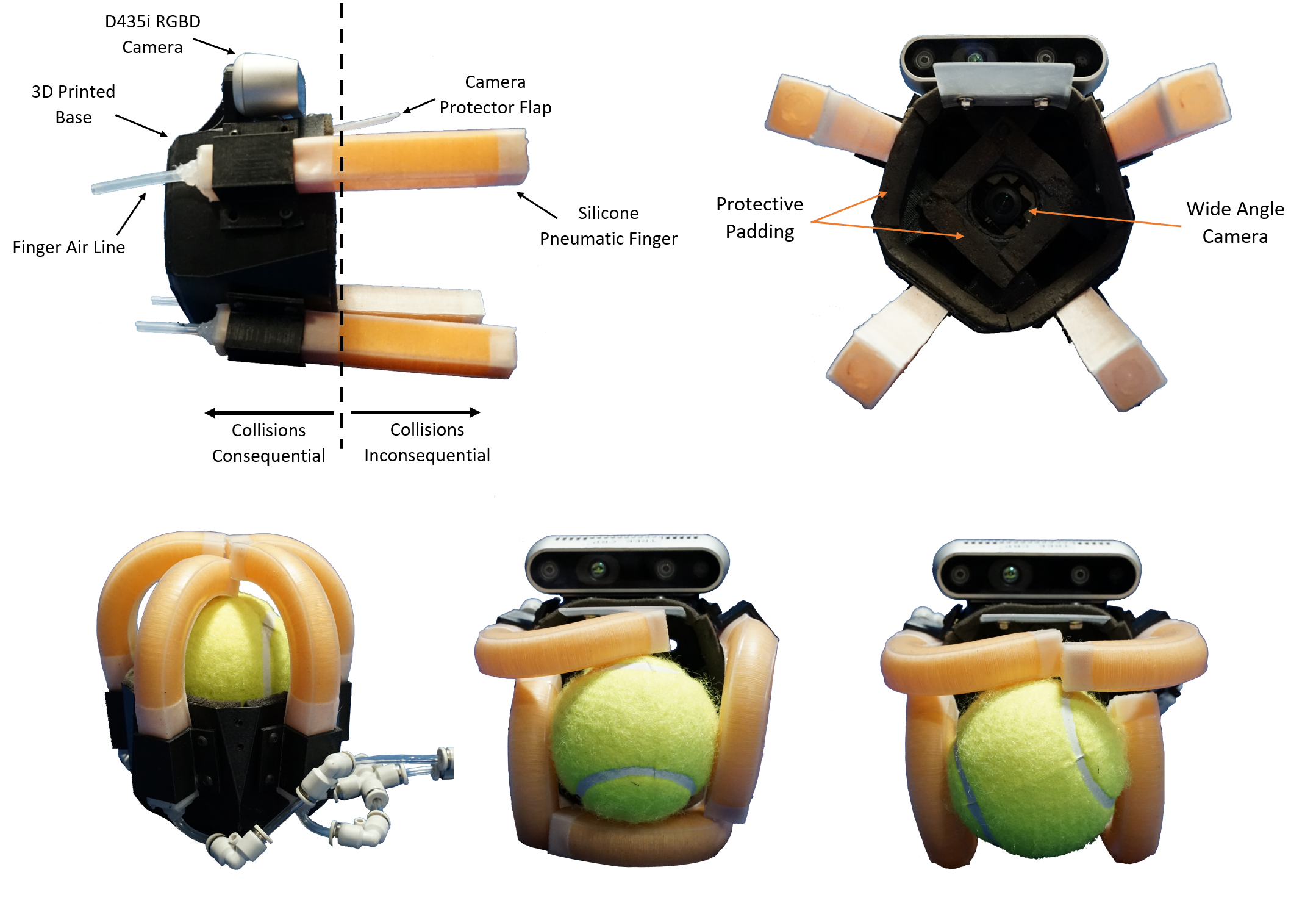}
	\caption{The soft robotic gripper showing the two zones of hard and soft components, as well as 3 commonly observed stable modes. Objects fully within the gripper cup usually resulted in the left most mode, while larger or further away objects would typically put the gripper in one of the other two stable modes shown. Different air pressures also impact gripper geometry and mode.}
	\label{fig:softGripper}
\end{figure}

Damaging contact with tree branches and fruit is a risk with any rigid gripper and will preclude the fully autonomous long term deployment of harvesting robots unless extremely reliable perception systems can be developed. By employing soft robotics components, key failure cases can be avoided in hardware rather than software. To demonstrate this, a second gripper was developed with 4 soft pneumatic fingers, shown in~Figure \ref{fig:softGripper}, using the designs from~\textcite{sun2017}, and~\textcite{sun2013}, all connected to a common air loop ensuring they close together. Control for this occurs on an Arduino Uno which actuates a pair of pneumatic solenoids to alternatively pressurise and depressurise the gripper air loop. By having only soft components extending beyond the gripper cup, collisions in this region can be ignored. 

	\section{Software and Algorithm Design}
\begin{figure}[h]
	\centering
	\includegraphics[width=0.75\columnwidth]{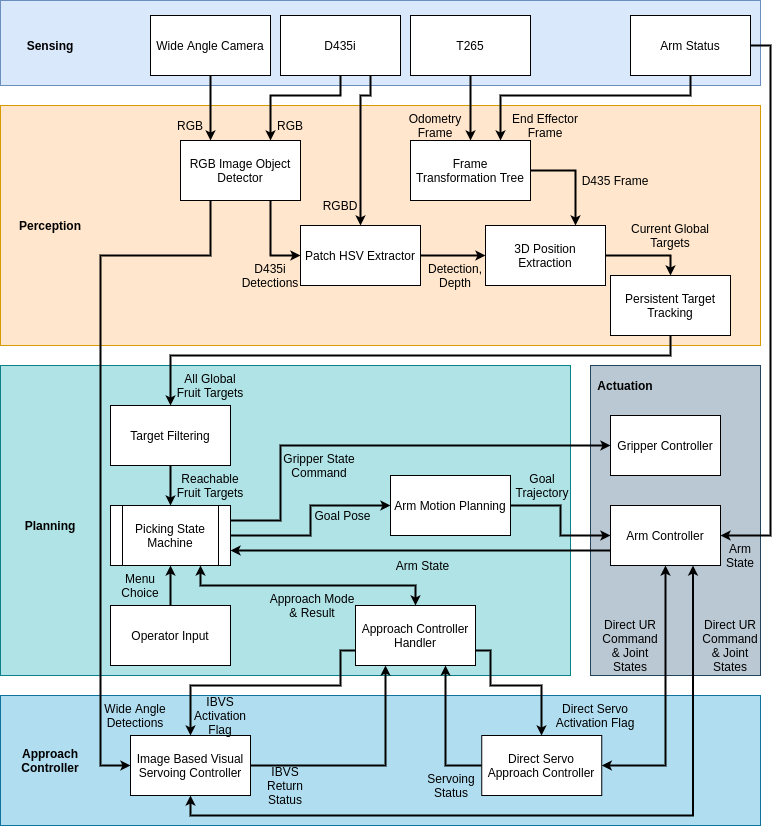}
	\caption{The key software modules. The Sensing block produces both image frames and transform frames which are processed by the Perception block into a coherent world representation of robot and fruit positions. The Planning block determines actuation motions, while the Approach Controllers provide low level feedback control of the actuation during final picking approach.}
	\label{fig:sw}
\end{figure}

The software architecture can be divided into 5 key modules, shown in Figure~\ref{fig:sw}. Sensing and perception form a standalone unit that can run without any additional input, allowing the same platform to be used for tree crop analytics such as fruit counting and yield volume estimation. The Robot Operating System (ROS) is used for message passing and software bringup. All key tuning values for the system are stored on the ROS parameter server and can be updated live without restarting any nodes. While this induced additional development overhead, the ability to live update all parameters was found to be key for field development. 

Dealing with hard and soft obstacles is a primary challenge of fruit harvesting and avoiding damaging collisions requires a combined hardware and software approach. In addition to soft robotics components, a two stage planning mechanism is used. First an approach pose 150mm out from the fruit location is planned to while respecting a 'hard obstacles' safety constraint plane just beyond that, this prevents the arm from moving through the trellis while reaching that pose. Following this, one of two final approach controllers takes over and moves the arm to the fruit location while not exceeding a second 'soft obstacles' safety plane which prevents the hard gripper components from contacting the trellis. With this two stage approach, only the soft gripper components can contact the trellis wires, tree limbs and posts. 

\begin{figure}[h]
	\centering
	\includegraphics[width=0.75\columnwidth]{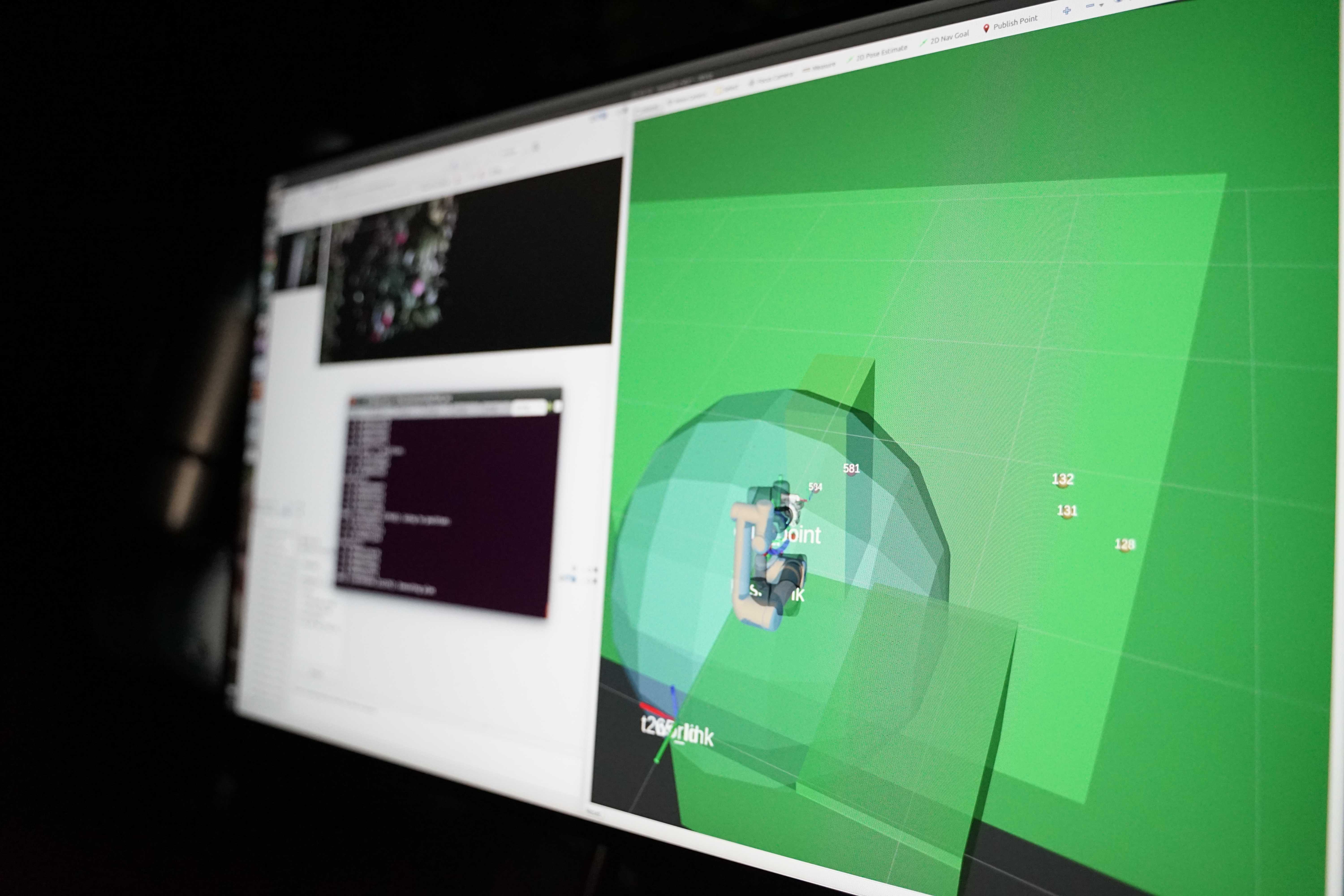}
	\caption{The visualisation output. Hard obstacles are shown in green, including the trellis plane (dark green background), approach plane (lighter green background) and trailer (nearest green box). The RoI is the rectangular green box in front of the arm. Targets are colour coded by their presence in the RoI or in the arm reachability sphere (shown in blue).}
	\label{fig:rviz}
\end{figure}

Visualisation occurs in Rviz, either onboard the generic PC or off-board via WiFi link. Shown in Figure~\ref{fig:rviz} is an example of the default visualisation window showing the arm state and planning scene obstacles. The arm reachability filter is shown as a blue sphere with the ROI as a green box. Target fruit are colour coded by their filtering status.  

\subsection{Sensing}
Framerates are limited to 15FPS for the D435i and wide angle camera, including the depth feeds, to reduce computation intesity. Overall CPU load for the system is approximately 60\%, of which 20\% is due to visualisation tools. The depth and image feeds to the patch Hue-Saturation-Value (HSV) extractor are likewise downscaled from 640x480 to 320x240 for efficiency. Odometry transformations from the T265 SLAM camera were found to be unstable under certain circumstances, so an option to disable this was also built in. All cameras operate with automatic exposure turned on.

\subsection{Perception}
Object detection is run on both the D435i and wide angle RGB streams by feeding these to a common ROS topic then splitting the output. A YoloV3 based architecture optimised for embedded hardware is deployed on the NVIDIA Xavier with image data being passed over the LAN. This reduces load on the primary computer and allows for very low power (30W) inference at over 15FPS. Detector training labels are constructed to capture the full target extents, even where these are obscured, this allows for better estimation of fruit size using just the bounding box. 

While deep learning based object detection does have much larger training and inference burdens than hand engineered features such as HSV thresholding, by using a commercial labelling service it was possible to gather data, label and train a model overnight for a 24hr turn around. Rapid detector training is important for crops that may change appearance each season. 

In order to deproject a D435i detection into the world frame, the pixel coordinates (u,v) are required along with a depth value (d). Calculation of the (d) value is done by extracting the image patch corresponding to the bounding box extents, then applying a HSV filter to that patch and masking only fruit coloured depth pixels. The median depth value is then returned. By using a patch which is known to contain a fruit, a highly tolerant HSV threshold filter can be used. In the case of no HSV thresholded pixels being present the depth value at the bounding box centroid is used. If all pixels fall within the threshold this approach reduces to the case of taking the bounding box median depth value. This masking technique is specifically required for instances where the fruit is more than 50\% obscured, otherwise the depth will correspond to the obscuring objects and be artificially low. For a spherical fruit with no obscurations, a maximum ratio of $\frac{\pi}{4}$ accurate bounding box depth pixels will come from the fruit. 

A 3D position extraction module takes the (u,v) bounding box centroid position and calculated (d) value, and uses the camera intrinsics and frame transformation tree to deproject each detection into the world frame. Checks on minimum and maximum depth and size are performed here. A motion filter is also applied to ignore detections while the camera is moving above a certain linear or angular velocity, this eliminates blurred detections. A fruit size property is set for each fruit object, which is the mean bounding box dimension projected to the depth value. Additional properties such as ripeness, pickability, health and visibility can also be set here to allow for future functionality. This module publishes a list of all currently detected fruit objects, which is used by the persistent target tracking module. 

This target tracking module is used to track all currently and previously seen fruit within the world frame. All observed fruit are assigned a unique ID and tracked within the state vector of an Extended Kalman Filter (EKF). For each currently observed fruit, a euclidean distance threshold is applied to associate it with an existing target. If there is no existing target, it is added to the state matrix, otherwise the EKF update step is run for re-observed targets. At each filter iteration a static state transition model is applied, though this could easily be modified to allow target motion models, such as for wind. When a current detector frame is received, all elements (tracked targets) in the EKF state matrix are projected into the current camera frame. If a target should be observed but is not, a counter is incremented for that target. At the end of each update loop, targets with counters above a set threshold are removed as false positive detections. 

The EKF state vector, default noise matrices and initial covariance matrix are given by

\begin{equation*}
state=\left[
\begin{array}{c}
x_1 \\
y_1 \\
z_1 \\
x_2 \\
y_2 \\
\vdots \\
z_n
\end{array}%
\right], \; \;
Q = 0.01\times\mathcal{I}_{n,n}, \; \;
R = 0.02\times\mathcal{I}_{n,n}, \; \;
P=\left[
\begin{array}{ccc}
0.05 & 0 & 0 \\
0 & 0.05 & 0 \\
0 & 0 & 0.1
\end{array}%
\right],
\end{equation*}%

where $\mathcal{I}$ is the identity matrix. A method for estimating these noise covariance matrices is presented in our previous work~\parencite{brown2019}. All units are in meters and the third element of $P$ is aligned with the camera depth axis (d) under our frame definitions, so has greater variance than the (u,v) pixel readings. An association distance of $0.03$m was used. 

\subsection{Planning}
Target filtering is the first planning step, this takes the tracking module output and applies a tag to fruit objects which are either within the arm workspace or within a more restrictive Region of Interest (RoI). By using a highly restrictive RoI it is possible to largely eliminate arm singularities within the workspace, essential for reliable functioning of the approach controllers. For all tests, picks were attempted within a $0.5\times 0.5 \times 0.8$m RoI window directly beside the trailer, with a mobile platform, this did not slow down the overall pick rate significantly. 

Core picking functionality is constructed as a state machine which controls the current arm goal, gripper actuation and approach controller status. Five poses are defined for the UR5 arm at any given time; a home position for convenience, a drop position where the fruit are deposited, a look position that gives a view of the entire RoI, a pick approach position 150mm offset from the fruit and the actual fruit position for harvesting. The first three of these are user defined, while the last is dynamically updated by the approach controller. Operator input consists of setting or moving to the above poses, manually actuating the gripper, listing current target poses or starting automatic picking. For automatic picking, the state machine caches the most recent list of tracked and filtered fruit, all targets within the RoI are then worked through in order of highest to lowest on the trellis. We note that many existing works have applied travelling-salesperson solutions or similar to harvest order planning. This is only beneficial if the drop position varies, in our case the total tour length only varies with the choice of first fruit target after leaving the home position. This may be optimally chosen as the fruit that projects furthest along the drop to home position vector. 

Once a target is popped from this list the arm is planned to the approach pose for that target and one of the two approach controllers is then activated. If the IBVS controller is used and fails, the direct approach controller is attempted. If both fail, likely due to a collision or dangerous singularity, the pick is aborted and the next target chosen. If either controller succeeds the gripper is closed and the arm moved backwards 150mm using the direct approach controller. From there, the arm is planned to the drop pose and the next target chosen. Feedback from the parallel gripper is used to determine when nothing has been contacted during a grab, in that case the IBVS controller is run again once from that position and a grasp reattempted.

Arm motion planning and Inverse Kinematics (IK) is handled using the RRTConnect planner from the Open Motion Planning Library~\parencite{sucan2012} and KDLKinematics IK solver within the Moveit ROS framework. This is run in threaded mode for concurrent planning and movement. Known paths, such as the drop to home motion, are precalculated and stored, these are also executed at a higher speed. For all tests a 1 second planning timeout was used, though this can be reduced in most cases. 

\begin{figure}[h]
	\centering
	\includegraphics[width=0.6\columnwidth]{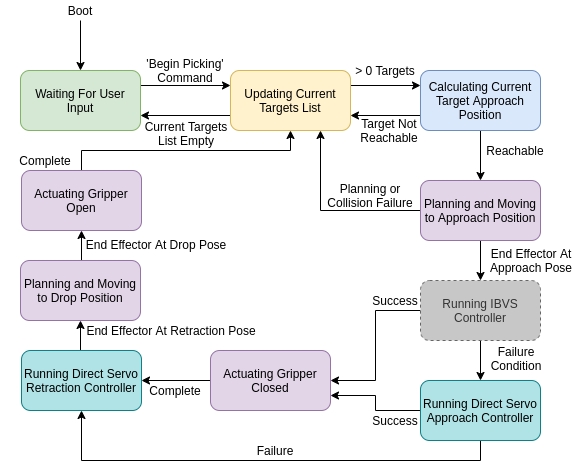}
	\caption{The picking state machine. Additional user input states, such as setting and recalling key arm positions are not shown. The IBVS controller was not found to be useful, as explained in Section~\ref{sec:cropObs}, so is included in the system design but was not used for field experiments}
	\label{fig:psm}
\end{figure}

\subsection{Approach Controller}
During final approach linear motion of the end effector is essential to avoid tree or hardware damage. One goal of this system is to assess the viability of a cartesian manipulator, so end effector orientation is restricted to be perpendicular to the trellis. Only cartesian motions are used during final approach, with the exception of rotational motion described in Section~\ref{sec:motion}.

The direct servo controller uses the UR script interface of the UR5 driver to send direct servo commands. This uses the instantaneous joint motion planner running on the UR control box and ensures the end effector moves exactly as desired. But software limitations prevent collision and singularity checking in this mode, thus the importance of a restricted RoI. Software joint limits are checked within both approach controllers by comparing current angles to the configured limits. If the joints are within 2 degrees of any limit, the controller is stopped and returns a failure condition. If the joints are over any limit, the previous velocity vector command is reversed and sent for one time step to bring them back within bounds, allowing the planner to work. 

Image based visual servoing is used to provide low level feedback control of the gripper final positioning. This was found to be essential in lab testing where contacting leaves and stems would cause the fruit to move during approach, it also makes the system robust to wind movement and perception errors. The nearest in-hand camera detection to the target point is tracked when the filter initialises, then subsequent frames attempt to associate a current detection with this target using a 2D distance. If association fails, the previous motion command is run again and a counter of missed frames is incremented. If this counter reaches a threshold, 15 by default, it triggers a failure status and the controller returns. Each successful association is used to calculate a pixel space error

\begin{equation*}
\delta_u = u_{detection} - u_{target}, \; \; \;
\delta_v = v_{detection} - v_{target}
\end{equation*}%

which has pixel space gains $G_u$ and $G_v$ applied to generate a velocity and is then used to scale the end effector depth axis velocity $d_{vel}$ against a preset value $ee_{vel}$

\begin{equation*}
u_{vel} = \delta_u G_u, \; \; \;
v_{vel} = \delta_v G_v
\end{equation*}%
\begin{equation*}
d_{vel} =  ee_{vel} - max(abs(u_{vel},v_{vel}))
\end{equation*}%

the three velocity components are combined into a normalised velocity vector which is scaled by the preset end effector velocity and then transformed from the camera optical frame to the arm controller frame. 

\begin{equation*}
\vec{V} = \left[
\begin{array}{c}
u_{vel} \\
v_{vel} \\
d_{vel}
\end{array}%
\right], \; \; \; 
\end{equation*}%

\begin{equation*}
\vec{V} = ||\vec{V}||_2 \times ee_{vel}
\end{equation*}%

While this results in a non-linear controller form, it was found to be stable and effective over a wide range of gains. The final velocity vector is sent directly to the arm controller as a commanded end effector motion. Detection size above a threshold is used as the successful stopping condition. Checks on failure conditions include singularities detected by high joint velocity, minimum and maximum bounding box sizes, traversal distance limit exceeded and controller timeout reached. 

\subsection{Actuation}
Direct arm actuation is handled by the UR controller box which implements PID joint control. Both velocity and torque limits can trigger a emergency stop on the arm, usually caused by a servoing singularity or collision respectively. These can be detected and reset in software, and cause the state machine to select the next target.  

Soft gripper actuation is binary and open loop. The parallel gripper provides torque sensing which is used to set a binary feedback flag indicating if something has been grasped. Both grippers are controlled using a ROS action service that sends serial commands via a common ASCII protocol to an Arduino for the soft gripper, or OpenCM9.04 for the parallel gripper. When a close command is received by the OpenCM9.04 it runs the servomotor until a torque or width limit is reached. The minimum closing width and torque both reside on the ROS parameter server and can be dynamically set using the global fruit object diameter if desired.

	\section{Testing Process}
The prototype system was first tested on an apple crop, to confirm all the components were working properly, before being evaluated on a commercial plum crop. Harvesting time windows in different crops make testing under identical conditions difficult, this is one motivating factor behind building a flexible and modular development platform. Some unexpected changes were required when moving to the plum crop, which are detailed in Section~\ref{sec:cropObs}.

System evaluation occurred on two rows of a commercial red plum cultivar known as 'Late Scheffer' grown in the 2D fruiting wall style of trellis. Fruit bunching was observed to be common, which also reduces the fruit sale quality where they are in contact. The grower indicated that he is trying to reduce bunching through more targeted thinning in the future, so this is not considered a major problem for harvest performance. Pick attempts were made on all detected fruit within the RoI box, the RoI height corresponds roughly to the space between the middle two trellis wires in all photos. After all fruit in the box were attempted, the platform was moved forward and stopped, leaving a 100mm horizontal RoI overlap with the last pick attempt. Fruit falling within this overlap are automatically excluded from a second pick attempt. No crop modifications or target exclusions were applied. Detected targets behind trunks, trellis wires and other fruit were still attempted. 

\label{sec:conditions}
System tuning and evaluation took place over a week under conditions of rain, high wind and darkness. The platform is able to operate in light rain by placing hook and loop secured flaps over the lower level and sealing the D435i. Wind caused a small amount of fruit movement, primarily in the depth axis which the gripper design is naturally tolerant of. Most fruit were within the typical picking ripeness window. Logs of all sensor feeds and system modules are made available for direct inspection of growing conditions\footnote{data.acfr.usyd.edu.au/Agriculture/PlumHarvesting}. 

\begin{figure}[h]
	\centering
	\includegraphics[width=0.6\columnwidth]{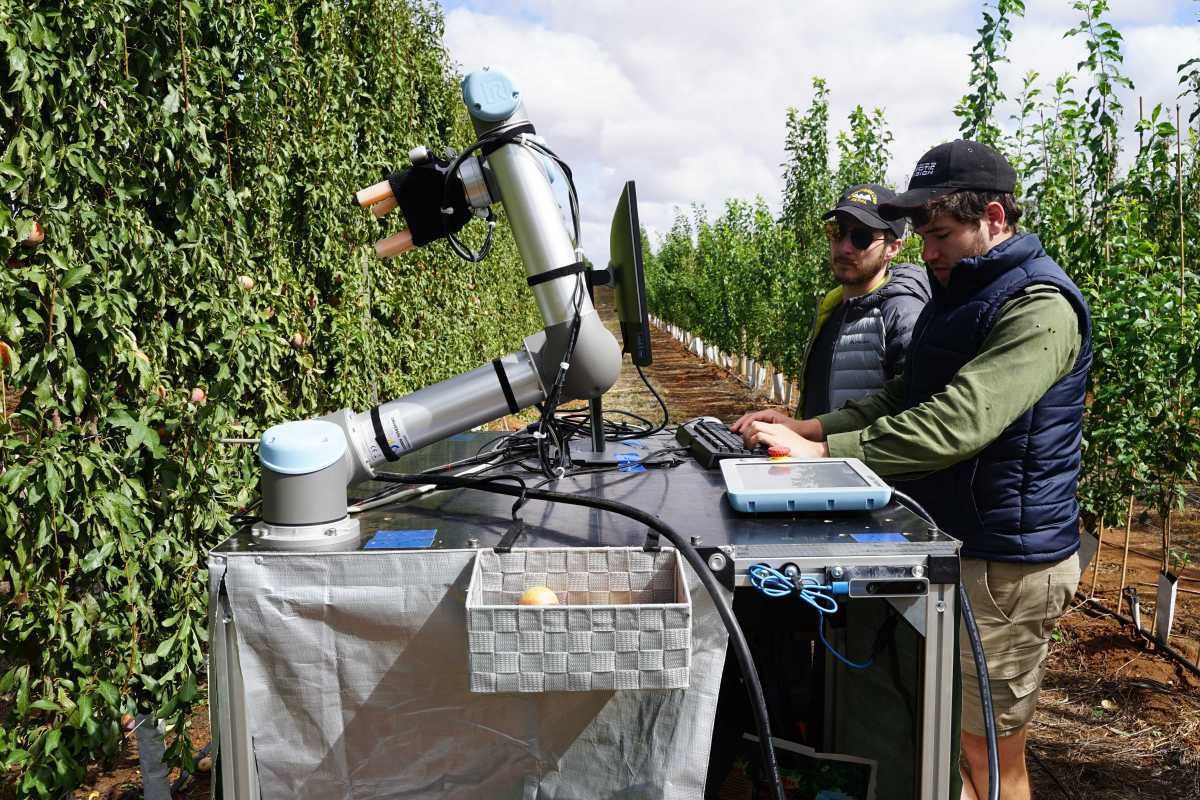}
	\caption{System during initial setup. Showing the user input terminal and arm control tablet on the right, soft gripper with skirt attached and T265 on the platform corner. Fruit outside the RoI had already been manually harvested for this section of trellis, no crop modification or picking was done within the RoI.}
	\label{fig:operation}
\end{figure}

Many different system architectures and algorithms have been suggested in the literature and without existing results on plums testing multiple module design choices was necessary. Specifically the choice of gripper, detector and picking motion were all assessed.

\subsection{Detector Type Study}
Effective object detection for harvesting is typically measured using recall and precision, which are essential for being able to harvest a large portion of the crop. However, commercial operation also places requirements on frame rates and power consumption, something not always achieved by high accuracy deep learning approaches. To compare a range of options, a basic HSV detector is compared to the low power embedded YoloV3 model and a state of the art Retinanet model. All of these are able to run at 15FPS. 

An HSV detector is built by manually tuning the 3 HSV channel thresholds on a representative daytime image to yield a binary mask of targets, which region size thresholds are applied to. Contiguous connected regions are considered a single object and returned as a positive detection. 

Both deep learning models are trained on 492 images, representing over 3000 bounding box examples from both day and night scenes. Default hyper-parameter values are used for these with training data augmentation turned on. The embedded YoloV3 model is tuned to preference very low false positive rates above high recall. The Retinanet model uses a Keras implementation\footnote{github.com/fizyr/keras-retinanet} running on the Zotac RTX2080, and a confidence cutoff of 0.7 to consider an object a true detection. 

Sixteen frames exhibiting no camera motion were randomly selected from both the day and night data. The true positive, false positive, false negative and mis-separation rates are calculated for these images. Mis-separation errors occur where a single bounding box has stretched to include multiple fruit, or multiple boxes exist for one fruit. Truncated bounding boxes and fruit are ignored. Bounding box accuracy was not found to be a limiting factor so is not further quantified. 

\subsection{Picking Motion Study}
\label{sec:motion}
Numerous complex motion strategies have been applied in existing works. Following discussions with human pickers and the grower, two motion strategies were proposed. The first is a simple straight-in, straight-out approach which is easily implemented on a cartesian system. The second was straight-in, rotate, angled-out which mimics the motion of human pickers, this is denoted 'complex motion'. 

\begin{figure}[h]
	\centering
	\includegraphics[width=0.7\columnwidth]{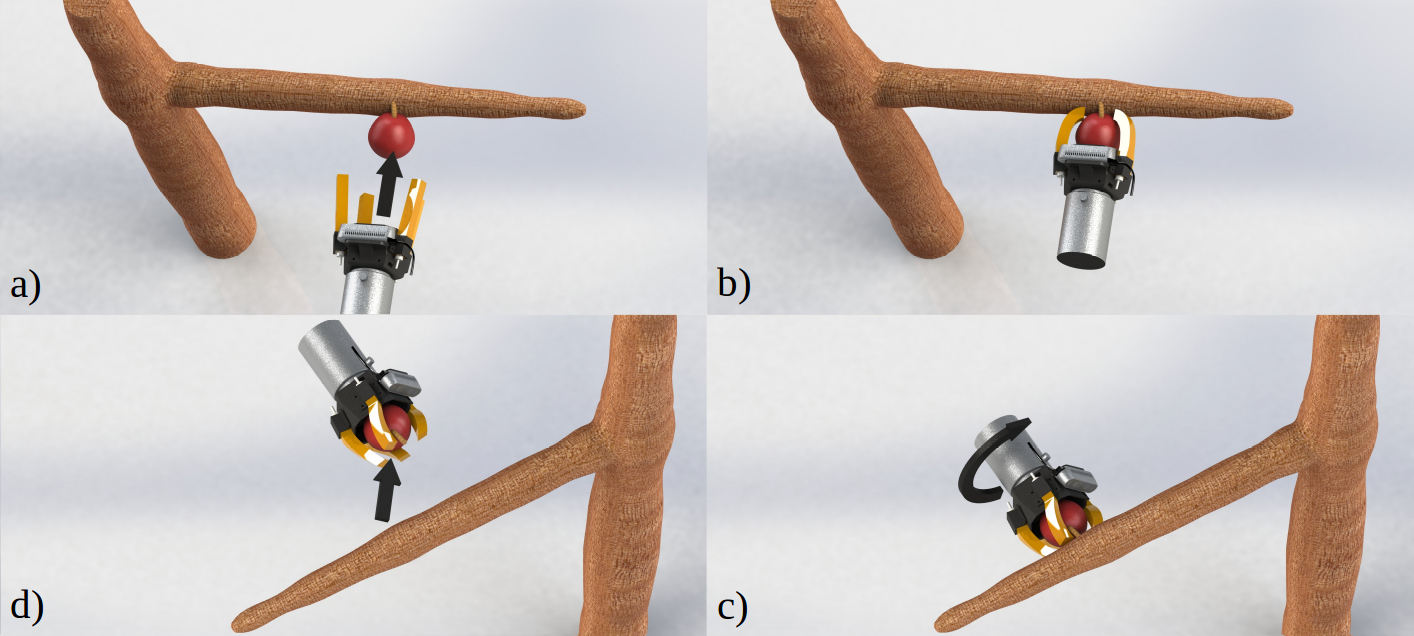}
	\caption{The complex picking motion. a) the gripper approaches the target fruit straight on, b) the fingers are actuated to grip the fruit, c) the gripper is twisted in place, d) the gripper is retracted at an angle to the trellis plane. }
	\label{fig:retractionAngles}
\end{figure}

\subsection{Gripper Type Study}
Use of soft components was primarily chosen to reduce the impact of collisions but will also alter harvesting performance. To assess this, the parallel gripper was manually positioned for a number of pick attempts in either a vertical or horizontal orientation. Rotational motion with this gripper was rarely possible due to collisions, so a straight-in, straight-out motion was used.

	\section{Results}
Overall harvesting success was far below that required for commercial viability, but represents a reasonable first step for this platform and numerous system components show clear opportunity for improvement. Time per pick was not a goal during testing because this is easily improved by using multiple concurrent actuators. The majority of time elapsed is due to actuation, and arm speed is constrained by worker safety and platform stability, rather than actuator limits. Lab testing, without the risk of damaging collisions, indicated a possible per pick time of 12 seconds, including all steps except trailer platform repositioning. 

\subsection{Detector Type}
\begin{figure}[h!]
	\centering
	\includegraphics[clip, trim=1.5cm 5cm 3cm 2cm, width=1.00\textwidth]{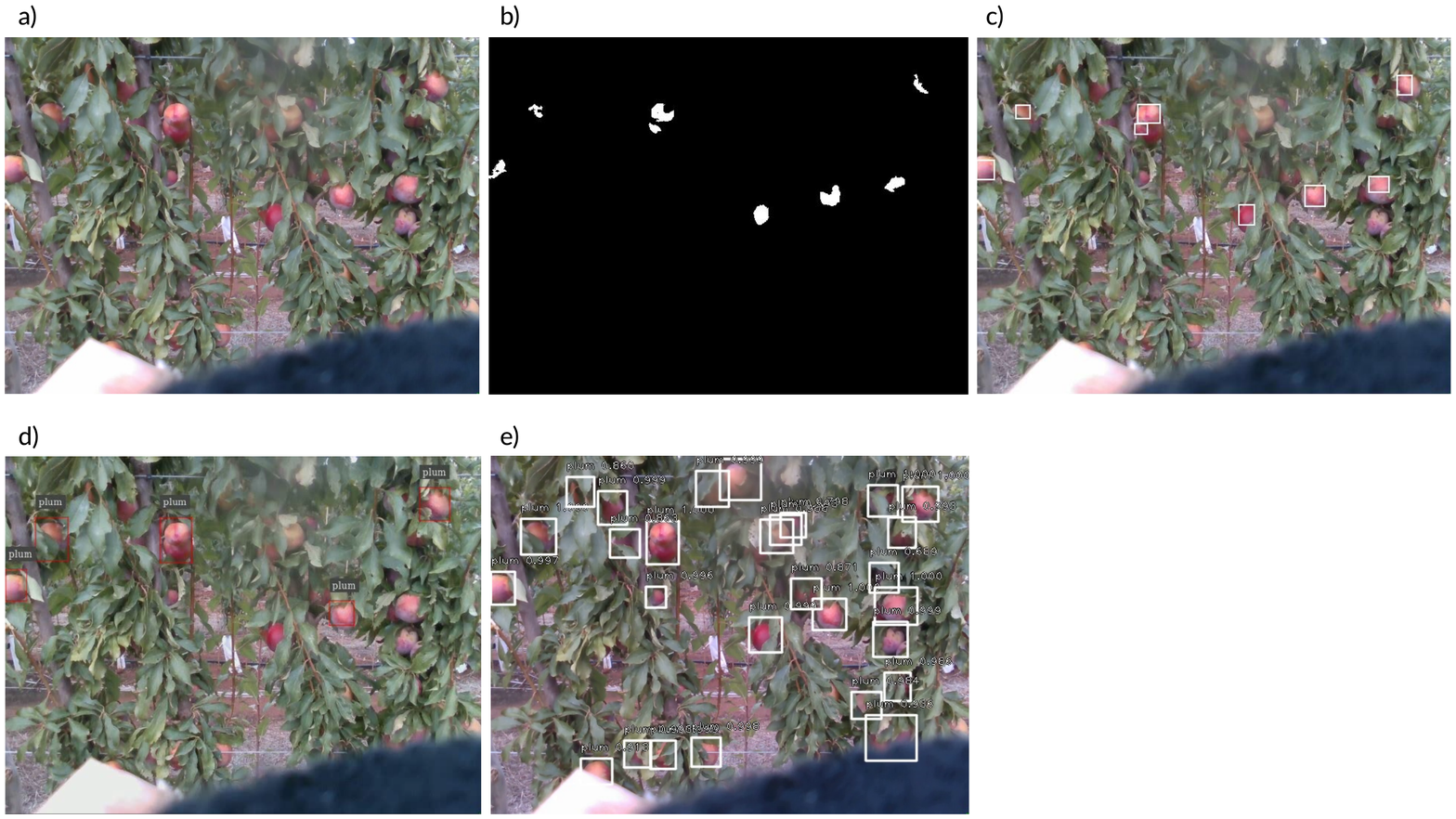}
	\caption{Detector comparison. a) the input frame, b) HSV mask, c) HSV detections, d) YoloV3 detections, e) Retinanet detections. The HSV detections image contains one bad box example, the rest are considered successful, though with inaccurate boxes. The left-most HSV detection would be ignored, due to truncation.}
	\label{fig:detectorComparison}
\end{figure}

HSV thresholding with region size filtering was found to be not at all effective. Although it performed well on red apples in the Sydney region during preliminary testing, the HSV channels in red plums could not be well separated from the red soil in the Victoria region, even under artificial light. The bounding boxes produced were also of lower quality, as in Figure~\ref{fig:detectorComparison}.

The embedded YoloV3 model achieved a very low false postive and mis-separation rate, at the cost of low recall. Practically the detection rate of this model is insufficient, even with the target tracking module able to compensate for detections which are missed in some frames. 

Table~\ref{tab:detectorComparison} shows the day and night performance of the 3 detectors. Moving to artificial lighting improved the HSV detector, despite it being tuned for daytime operation. Retinanet also performed slightly better at night while YoloV3 was less effective. 

\begin{table}[h!]
	\centering
	\begin{tabular}{lccccccc}
		\textbf{Detector} & \multicolumn{1}{l}{\textbf{\begin{tabular}[c]{@{}c@{}}True \\ Positives\end{tabular}}} & \multicolumn{1}{l}{\textbf{\begin{tabular}[c]{@{}c@{}}False \\ Positives\end{tabular}}} & \multicolumn{1}{l}{\textbf{\begin{tabular}[c]{@{}c@{}}False \\ Negatives\end{tabular}}} & \multicolumn{1}{l}{\textbf{\begin{tabular}[c]{@{}c@{}}Mis- \\ Separation\end{tabular}}} & \multicolumn{1}{l}{\textbf{\begin{tabular}[c]{@{}c@{}}Recall\end{tabular}}} &
		\multicolumn{1}{l}{\textbf{\begin{tabular}[c]{@{}c@{}}Precision\end{tabular}}} &
		\multicolumn{1}{l}{\textbf{\begin{tabular}[c]{@{}c@{}}Bad Box \\ Rate\end{tabular}}}   \\ \hline
		HSV Day & 57 & 14 & 170 & 24 & 25.1\% & 80.3\% & 42.1\% \\
		YoloV3 Day  & 73 & 0 & 154 & 2 & 32.2\% & 100.0\% & 2.7\% \\  
		Retinanet Day & 185 & 7 & 42 & 8 & 81.5\% & 96.4\% & 4.3\% \\
		HSV Night & 30 & 6 & 40 & 4 & 42.9\% & 83.3\% & 13.3\% \\
		YoloV3 Night  & 13 & 0 & 57 & 1 & 18.6\% & 100.0\% & 7.7\% \\
		Retinanet Night & 61 & 0 & 9 & 4 & 87.1\% & 100.0\% & 6.6\%
	\end{tabular}
	\caption{Detector performance table}
	\label{tab:detectorComparison}
\end{table}

Retinanet significantly outperformed the embedded YoloV3 model but recorded an average GPU power draw of 168W, over four times that of the Xavier embedded computer. The low recall rate of YoloV3 was unexpected and may be due to the conversion process to an embedded model type, a more detailed comparative study of current deep learning object detectors for fruit will be performed in future work.

\subsection{Picking Motion}

Soft gripper failure modes and rates are enumerated in Table~\ref{tab:softFailures}. The simple straight-in, straight-out motion was not very effective and frequently resulted in insufficient gripper force to detach the fruit. There are two reasons for this, the first being that only a subset of the fingers would frequently be in contact with the fruit leading to lower overall force transferred to the target. The second is due to fruit stems having much higher tensile strength, as opposed to shear strength. Mature plums detach at the abscission layer, which is the natural process by which they fall from the tree. However, pulling directly along the stem axis often resulted in stem pull-out which damages the fruit and requires a much higher amount of grip force. Rotational motion while pulling, or applying force at an angle to the stem, resulted in more detachments occurring at the abscission layer with lower required force.  By applying the rotation and angled pull back motion, the fingers also fall into a more tightly closed stable mode, shown in Figure~\ref{fig:softGripper} bottom left, where the target is fully within the rigid gripper cup. 

\begin{table}[]
	\centering
	\begin{tabular}{lcccc}
		\textbf{Outcome}                                                     & \textbf{Straight} & \textbf{Angled} & \textbf{\begin{tabular}[c]{@{}l@{}}Straight\\ Percentage\end{tabular}} & \textbf{\begin{tabular}[c]{@{}l@{}}Angled\\ Percentage\end{tabular}} \\ \hline
		Success                                                              & 4                 & 27              & 20.0\%                                                                 & 42.2\%                                                               \\
		\begin{tabular}[c]{@{}l@{}}Grip Force\\ Failure\end{tabular}         & 7                 & 5               & 35.0\%                                                                 & 7.8\%                                                                \\
		\begin{tabular}[c]{@{}l@{}}Bad Positioning\\ Failure\end{tabular}    & 6                 & 9               & 30.0\%                                                                 & 14.0\%                                                               \\
		\begin{tabular}[c]{@{}l@{}}Knocked Off\\ Target Failure\end{tabular} & 1                 & 9               & 5.0\%                                                                  & 14.0\%                                                               \\
		Gripper Failure                                                      & 0                 & 2               & 0.0\%                                                                  & 3.1\%                                                                \\
		Other Failure                                                        & 2                 & 12              & 10.0\%                                                                 & 18.8\%                                                               \\
		Total                                                                & 20                & 64              &                                                                        &                                                                     
	\end{tabular}
	\caption{Gripper performance and failure modes by motion type. Straight and angled percentages refer to the relative rate of that outcome for the given motion type.}
	\label{tab:softFailures}
\end{table}

Bad positioning refers to any case where it was judged that more accurate positioning may have led to a successful pick. Often this was the result of fruit falling through the finger gaps in the gripper as it retracted. Stated failure modes are overlapping and must be considered in order, for instance a greater grip force may correct for a badly positioned pick. Likewise an attempt that fails due to positioning may also have subsequently failed due to grip force even with perfect positioning.

A large portion of the 'other' failure types for the soft gripper were fruit in difficult positions, such as on the other side of a trellis wire or trunk. Use of the more complex motion also resulted in many more fruit being knocked off, typically a missed pick during simple motion would leave the target fruit on the tree. When gripper rotation is introduced the fruit are much more effectively detached, but if not well gripped will then fall to the ground. 

\subsection{Gripper Type}
Both gripper type and retraction motion were found to have a large effect on picking success rate. Specifically, the hard parallel gripper effectiveness tripled when used in a vertical rather than horizontal orientation. It was observed that more fruit have hard obstacles on their sides than above or below them and the majority of hard gripper failures were a result of grasping with an obstacle between the finger and fruit.

\begin{table}[]
	\centering
	\begin{tabular}{lccc}
		\textbf{Manipulator Type}                                              & \multicolumn{1}{l}{\textbf{Successes}} & \multicolumn{1}{l}{\textbf{Failures}} & \multicolumn{1}{l}{\textbf{\begin{tabular}[c]{@{}l@{}}Success\\ Rate\end{tabular}}} \\ \hline
		\begin{tabular}[c]{@{}l@{}}Soft Gripper \\ Simple Motion\end{tabular}  & 4                                      & 16                                    & 20.0\%                                                                              \\
		\begin{tabular}[c]{@{}l@{}}Soft Gripper \\ Complex Motion\end{tabular} & 27                                     & 37                                    & 42.2\%                                                                              \\
		\begin{tabular}[c]{@{}l@{}}Horizontal \\ Parallel Gripper\end{tabular} & 2                                      & 8                                     & 10.0\%                                                                              \\
		\begin{tabular}[c]{@{}l@{}}Vertical \\ Parallel Gripper\end{tabular}   & 6                                      & 7                                     & 30.0\%                                                                             
	\end{tabular}
	\caption{Gripper comparison table}
	\label{tab:gripperComparison}
\end{table}

No fruit damage was observed with either gripper, although the parallel gripper resulted in minor tree damage where collisions occurred. Zero emergency stops due to collisions were registered with the soft gripper in place. 
	\section{Discussion}
Overall goals of testing system modularity, module choice comparisons and evaluation on a plum crop were met. Picking performance was lower than anticipated but comparisons to existing work will not be very informative for a range of reasons, including the crop type and testing conditions. The modularity criteria was well met, with minimal issues when swapping over components for module testing. 

Several lessons around system design and development were learned. One clear lesson was the importance of rigorous and frequent testing. This is made difficult by the short harvesting window for most crops, uniqueness of each plant and trellis style, and the difficulty of creating accurate lab based testing setups. Many existing harvesting systems use a visual servoing approach controller, whereas this was found to be worse than going directly to the estimated fruit position for plum crops. This led to wasted development effort and unnecessary sensing hardware, which could have been reduced with earlier testing on plums.

Target tracking and filtering is an essential tool if less than perfect object detectors are used, such as in highly obscured crops. This not only compensates for some detector failures, but allows for additional fruit properties such as ripeness, to be estimated from multiple sensor frames or modalities. 

Compute capacity of the current system was sufficient but not excessive. Utilising an embedded solution for deep object detection saved energy while providing a stable and predictable frame rate. Bandwidth over all system topics was approximately 40 MB/s and the current software is not optimised for efficiency. This demonstrates that computing requirements can be easily met with commercial systems and off the shelf software frameworks, such as ROS.

Despite the water resistant properties of the platform, heavy rain caused the loss of several testing days. While building waterproof hardware is not difficult, the performance of most perception and gripping systems remains untested for heavy rain and could pose unforeseen issues. Night operation, under artificial light was successful with little drop in detector or depth accuracy. Human pickers typically harvest at night due to better weather conditions and lower wind, day harvesting of this cultivar is not possible during very hot weather as the fruit bruise when stacked in collector bins. 

The key software assumptions around soft and hard obstacle planning planes proved to be correct. With no consequential collisions occurring in the soft obstacle area and no hard components colliding with the trellis or trunk. This was in part due to careful calibration of these planes, which could in future be done online using sensor data, and also due to the uniform and well kept trellis structure. We stress that no modifications were made to the crop for this research work, it is a commercial crop undergoing an otherwise normal harvest process. Fruiting wall plum crops do appear to be a slightly more difficult, or at least different, problem than apple harvesting. Primarily due to fruit proximity to branches and trunks, though thinning protocol also impacts this. 

\subsection{Crop Specific Observations}
\label{sec:cropObs}
When the prototype system was previously tested on apples it performed subjectively well. The additional size and separation of these from the growing trunk led to the current soft gripper design. HSV detection also worked very well on red apples and the bunching was less severe, leading to less collateral damage fruit. The evaluation plum crop grows extremely close to woody trunks and branches, while being a lot smaller in size. This meant the soft gripper was oversized and needed to be modified with a soft skirt to prevent fruit from escaping between the fingers. While growing systems will vary greatly, even within plum crops, the proximity of target plums to branches caused many collisions and made the use of an angled motion critical to pick success. Images of the growing conditions can be found in the online data logs.

While the visual servoing approach controller was not originally intended to be one of the module design choice tests, it quickly became apparent that it was not contributing to picking success. Lab testing on fake orange trees and pre-evaluation tests on apples had indicated that the IBVS controller was essential to compensating for fruit motion. Unlike these crops, the tested plum crop has extremely short stems and exhibited almost no motion during picking, making the IBVS controller counter productive, even with the very small number of failures attributable to that module.

Some fruit and market combinations require the stem to be present, while for others it must be removed. In this plum crop either was acceptable, but stem removal often led to stem pull out where the plum skin is broken, rendering it worthless. Where stems are required for commercial sale, the picking motion strategy must reflect this. 

Crop style, gripper design and motion choice are all closely interlinked. These can't be separated and will pose a major problem for future development. The seasonal availability of testing crops compounds these difficulties, though some work has been done to address this using simulation tools~\parencite{wang2018}. 

\subsection{Module Design Choices}
Choice of detector architecture had a surprising impact on performance, with a significant difference in precision and recall between the two deep networks. The optimal balance of these two objectives will depend on the cost of attempting bad picks due to false positives, weighed against the missed picks due to false negatives. Detector limitations were partially compensated for by the persistent target tracking and filtering. The detector type study is intended to be comparative only and careful further tuning could marginally improve the performance of all 3 detector types.

Picking motion was also important, with even minor changes to the soft gripper motion resulting in different pick success figures. This does come at the cost of increased actuation time and additional gripper wear and tear. Longevity was one major downside of the soft gripper, with 2 catastrophic and 2 minor finger failures observed even for the small number of picks, approximately 300, carried out over the week. All motion strategies resulted in some fingers being caught between hard components and tree branches leading to external fabric and internal structure damage. Contact with rough old growth wood caused abrasions to the external silicone, introducing failure points when this was deformed by the finger actuation. More robust material selection should significantly improve soft gripper lifespan.  

While the parallel gripper vertical performance was not vastly worse than the soft gripper, it resulted in many emergency stops due to detected collisions. The small number of attempted picks is due to this. Collision emergency stops are not captured in the success rate figures but makes the deployment of a hard gripper infeasible without much more advanced perception algorithms. Some of our previous work~\parencite{hung2013}, has addressed trunk detection in trellises but locating obscured branches is a difficult open problem.

\section{Conclusions}
In this work we lay out the design and evaluation of a modular research platform for robotic plum harvesting. The growing conditions, primarily fruit proximity to branches, led to unanticipated results and a lower than expected success rate highlighting the importance of testing on additional tree crops to those presently in the literature. Visual servoing was found to be unnecessary, but target tracking and filtering was essential. Small changes to picking motion had critical impacts on harvesting success and must be considered in the context of gripper design.

Growers in Australia and overseas have demonstrated their willingness to adopt growing systems suitable for mechanisation and eventual automation. Fruit walls appear to be a promising candidate based on this work and current results on apples. But flower thinning practices are also essential to crop success and are rarely assessed in harvesting tests, beyond the impact of bunched fruit, which remains a major problem for robotic harvesters.  

Soft robotics has shown good results in building effective harvest systems that minimise fruit, tree and robot damage resulting from collisions. Long term reliability is an issue for the design presented here and will require thorough reliability testing for any commercial systems seeking to use this technology. 

Next year we hope to return to the same crop to compare inter-year performance and further develop system modules specific to plum harvesting. The hard-and-soft collision distinction used here for the soft gripper, goes a long way towards addressing collisions, but a unified framework for detecting, categorising and recovering from all collision cases in robotic harvesting is still lacking. We also hope to improve soft gripper force and robustness in further design iterations. An open problem of this system and others, is autonomous detection of difficult fruit which should be avoided in picking attempts.

	\vspace*{\fill}
	
	\subsubsection*{Acknowledgements}
	The authors would like to thank Tom Ingram, Eric Dreischerf and Khalid Rafique for their tireless help on this project. Also Yi Sun for his soft robotics expertise and assistance.
	
	\newpage
	
	\printbibliography
	
	\newpage
	
	\theendnotes
	
\end{document}